\pdfoutput=1

\documentclass[11pt]{article}

\usepackage[final]{acl}

\usepackage{times}
\usepackage{latexsym}

\usepackage[T1]{fontenc}

\usepackage[utf8]{inputenc}

\usepackage{microtype}

\definecolor{thedarkblue}{RGB}{0,0,120} %
\definecolor{mydarkblue}{rgb}{0,0.08,0.45} %
\definecolor{darkbluenew}{rgb}{0,0.08,180}
\usepackage{hyperref}
\hypersetup{%
    colorlinks=true,
    linkcolor=mydarkblue,
    citecolor=mydarkblue,
    filecolor=mydarkblue,
    urlcolor=mydarkblue}

\usepackage{inconsolata}

\usepackage[strict]{changepage}

\usepackage{framed}

\definecolor{formalshade}{rgb}{0.95,0.95,1}

\usepackage{rotate}
\usepackage{adjustbox}
\usepackage{array}
\usepackage{capt-of}
\usepackage{tabulary}
\usepackage{setspace}
\usepackage{amssymb}
\usepackage{mathtools}
\usepackage{pifont} 
\usepackage{go} %

\definecolor{figblue}{RGB}{0,176,240}
\definecolor{figgreen}{RGB}{78,173,91}

\usepackage{colortbl}
\usepackage{xcolor}
\usepackage{graphicx}
\usepackage{booktabs}
\usepackage{array}

\usepackage{amsmath}
\usepackage{amssymb}
\usepackage{mathtools}
\usepackage{amsthm}
\usepackage{algorithm}
\usepackage[noend]{algpseudocode}

\usepackage{xcolor}
\usepackage{colortbl}
\definecolor{thedarkblue}{RGB}{0,0,120} 
\definecolor{mydarkblue}{rgb}{0,0.08,0.45} 
\definecolor{darkblue}{rgb}{0,0.08,180}
\colorlet{TufteRed}{red!80!black}
\definecolor{theblue}{RGB}{0,0,180}
\colorlet{thered}{TufteRed}
\usepackage{xcolor}

\usepackage{microtype}
\usepackage{balance}

\usepackage{booktabs}
\usepackage{tabularx}

\usepackage{amsmath,amssymb,amsthm}

\newcommand{\eat}[1]{\ignorespaces}
\usepackage{comment}

\newcommand{\journal}[1]{} 

\usepackage{tikz}
\usepackage{verbatim}
\usetikzlibrary{arrows}
\usetikzlibrary{shapes,snakes}
\usetikzlibrary{decorations.pathmorphing} 
\usetikzlibrary{fit}					
\usetikzlibrary{backgrounds}	

\usepackage{ragged2e}
\usepackage{multirow}
\usepackage{microtype}
\usepackage{balance}
\usepackage{setspace}

\graphicspath{{./}{./graphics/}}
\newcolumntype{H}{>{\setbox0=\hbox\bgroup}c<{\egroup}@{}}

\newcolumntype{R}[1]{>{\RaggedLeft\arraybackslash}} 
\newcolumntype{L}[1]{>{\RaggedRight\arraybackslash}} 

\newcommand\TTT{\rule{0pt}{3.2ex}}

\newcommand{\eg}{\emph{e.g.}}

\AtBeginEnvironment{pmatrix}{\setlength{\arraycolsep}{2pt}}

\renewcommand{\vec}[1]{\boldsymbol{\mathrm{#1}}}

\DeclareMathOperator{\hugeE}{\mbox{\huge\raise-0.3ex\hbox{E}}}
\DeclareMathOperator{\p}{\mathbb{P}}
\DeclareMathOperator{\hugep}{\mbox{\huge\raise-0.3ex\hbox{$\p$}}}

\providecommand{\vy}{\ensuremath{\vec{y}}}

\usepackage{subfigure}
\usepackage{nicefrac}

\DeclareMathAlphabet{\mathbcal}{OMS}{cmsy}{b}{n}
\usepackage{amsmath}
\usepackage{mathrsfs}
\usepackage{comment}

\usepackage{bm}
\usepackage{bbm}
\usepackage{amssymb}
\usepackage{enumitem}
\usepackage{paralist}
\usepackage{cleveref}

\title{Exploring Rewriting Approaches for Different Conversational Tasks}

\author{Md Mehrab Tanjim, Ryan A. Rossi, Mike Rimer, Xiang Chen, Sungchul Kim,\\ 
\textbf{Vaishnavi Muppala, Tong Yu, Zhengmian Hu, Ritwik Sinha} \\
\textbf{Wei Zhang,  Iftikhar Ahamath Burhanuddin, Franck Dernoncourt}
\\
\{tanjim, ryrossi, mrimer, xiangche, sukim, mvaishna, tyu, zhengmianh, \\ risinha, wzhang, burhanud, dernonco\}@adobe.com
\\
  Adobe Research \\
  }

\begin{document}
\maketitle

\begin{abstract}
Conversational assistants often require a question rewriting algorithm that leverages a subset of past interactions to provide a more meaningful (accurate) answer to the user's question or request. However, the exact rewriting approach may often depend on the use case and application-specific tasks supported by the conversational assistant, among other constraints. In this paper, we systematically investigate two different approaches, denoted as rewriting and fusion, on two fundamentally different generation tasks, including a text-to-text generation task and a multimodal generative task that takes as input text and generates a visualization or data table that answers the user's question. Our results indicate that the specific rewriting or fusion approach highly depends on the underlying use case and generative task. In particular, we find that for a conversational question-answering assistant, the query rewriting approach performs best, whereas for a data analysis assistant that generates visualizations and data tables based on the user's conversation with the assistant, the fusion approach works best. Notably, we explore two datasets for the data analysis assistant use case, for short and long conversations, and we find that query fusion always performs better, whereas for the conversational text-based question-answering, the query rewrite approach performs best.
\end{abstract}

\section{Introduction}
Conversational assistants have become integral tools in various domains, from customer service to personal assistance. These systems often depend on sophisticated algorithms to interpret and respond to user queries effectively. One critical aspect of such systems is question or query rewriting (QR), which aims to transform user queries into more detailed and self-contained versions, thereby enhancing the system's ability to provide accurate responses.  

QR includes various techniques aimed at addressing underspecified and ambiguous queries. Traditionally, QR involves adding terms to the original query, known as query expansion \cite{lavrenko2017relevance}, rephrasing the query with similar phrases \cite{zukerman2002lexical}, or using synonymous terms \cite{jones2006generating}. Recently, the emergence of large language models (LLMs) has generated interest in utilizing these generative models to automatically resolve ambiguities during query processing, thereby enhancing query modeling for downstream tasks. For example, recent studies have prompted LLMs to generate detailed information, such as expected documents or pseudo-answers \cite{wang2023query2doc, jagerman2023query, anand2023context}. These methods are particularly useful when a high-quality dataset for a domain is unavailable, necessitating the employment of 
LLMs customized for the particular use-case. 

This leads to a crucial research problem: Can a single LLM-based query rewrite module be universally effective across diverse conversational scenarios, or is there a need for specialized modules tailored to specific query types and use cases? Addressing this problem involves systematically evaluating the performance of LLMs in query rewriting across various domains and tasks, identifying potential limitations, and exploring whether a one-size-fits-all approach is feasible or if a more nuanced, context-dependent strategy is required. This investigation will provide valuable insights into the adaptability of LLMs and inform the design of more robust and versatile conversational assistants.  

\medskip\noindent\textbf{Summary of Main Contributions.} 
The key contributions of this work are as follows:

\begin{compactitem}
    \item We introduce a generalized framework for two query rewriting approaches that is expressive to recover both approaches at either extreme.

    \item We systematically investigate both approaches across three new datasets including both short and long conversational data analysis task for generating visualizations from text and another text-based question-answering task.
    
    \item We find that for both the short and long conversational text-to-vis tasks, query fusion performs the best whereas for conversational question-answering task, the query rewrite outperforms query fusion.
    
\end{compactitem}

\begin{table*}[h!]
\centering
\small
\caption{Conversation that a user may have with an assistant and the corresponding rewritten question from our query fusion in Sec.~\ref{sec:query-fusion} with $k=1$.
}
\vspace{-2mm}
\label{table:conv-and-rewritten-question-query-fusion}
\begin{tabular}{%
@{} 
c
p{4.8cm} p{10cm}}
\toprule
& 
\textbf{Conversation} & \textbf{Rewritten Question} \\
\midrule
1 & compare monthly revenue by country & compare monthly revenue by country \\
2 & yearly & compare yearly revenue by country \\
3 & show it as a line chart & compare yearly revenue by country as line chart \\
4 & now change to marketing channel & compare yearly revenue by marketing channel as line chart \\
5 & what about month over month as bar & compare month over month revenue by marketing channel as bar \\
6 & replace with pageviews & compare month over month pageviews by marketing channel as bar \\
7 & show top-3 & compare month over month pageviews by top-3 marketing channels as bar \\
8 & what about top-5 & compare month over month pageviews by top-5 marketing channels as bar \\
9 & show only this month & compare this month pageviews by top-5 marketing channels as bar  \\
10 & add revenue & compare this month pageviews and revenue by top-5 marketing channels as bar  \\
\bottomrule
\end{tabular}
\end{table*}

\section{Related Work} \label{sec:related-work}

Query rewrite or reformulation (QR) includes various techniques aimed at addressing underspecified and ambiguous queries by transforming them into more detailed and self-contained versions that are appropriate for retrieval or question-answering systems. Traditionally, QR involves adding terms to the original query, known as query expansion \cite{lavrenko2017relevance}, rephrasing the query with similar phrases \cite{zukerman2002lexical}, or using synonymous terms \cite{jones2006generating}. When human-generated rewrites or reward signals are available, language models (LMs) are also trained specifically for question rewriting (QR) \cite{elgohary2019can, anantha2021open, vakulenko2021question, qian2022explicit, ma2023query}. In conversational search, query reformulation is employed to manage conversational dependencies. \cite{anantha2021open} introduce the rewrite-then-retrieve pipeline, which relies on a human-crafted dataset \cite{elgohary2019can}. Many studies fine-tune QR models to generate standalone questions \cite{yu2020few, voskarides2020query, lin2021multi, kumar2020making, wu2022conqrr}.

Recently, the emergence of large language models (LLMs) has generated interest in utilizing these generative models to automatically resolve ambiguities during query processing, thereby enhancing query modeling for downstream tasks. These tasks frequently aim to improve information retrieval in a single question answering setting. For example, recent studies have prompted LLMs to generate detailed information, such as expected documents or pseudo-answers \cite{wang2023query2doc, jagerman2023query, anand2023context}. These methods are particularly useful when a high-quality dataset for a specific domain is unavailable, necessitating the employment of off-the-shelf LLMs customized for the particular use-case. 

While these techniques significantly mitigate the issue of the extensive training data required for dedicated model training, prompting large language models (LLMs) for query rewriting introduces its own set of challenges. For instance, \citet{anand2023context} identified that LLM-based query rewriting can experience concept drift, deviating from the original intention or meaning when queries alone are used as prompts. This issue becomes particularly pronounced in real-world enterprise systems, where generic terms can have specific, context-dependent meanings. Our observations confirm this phenomenon (e.g., ``people" might refer to a specific metric or dimension in a chart, while ``segment" could denote a particular data object within the system), leading to the conclusion that there is no universal solution applicable to all use cases. Instead, practitioners must carefully consider their specific use case,
identifying the taxonomy and nature of the queries to design the most effective rewrite strategies. To this end, we present best practices for rewrite strategies and, through two distinct use cases within the same enterprise setting, demonstrate how these practices lead to optimal results.

\section{Approaches} \label{sec:approach}
In this section, we introduce a general parameterized approach in Algorithm~\ref{alg:query-rewrite} for query rewriting tasks.
First, we provide an overview of the parameterized query rewrite approach (Sec.~\ref{sec:overview}), then discuss two different parameterizations of it for text-based Q\&A rewrite (Sec.~\ref{sec:query-rewrite}) and query fusion task (Sec.~\ref{sec:query-fusion}).

\subsection{Overview} \label{sec:overview}
In Algorithm~\ref{alg:query-rewrite}, we first initialize $C$ to be empty, which will be used to represent the context given to the model (Line~\ref{algline:init}).
Next, we iteratively construct the appropriate context $C$ with the last $k$ previous input questions from $\mathcal{I}=\{I_1,\ldots,I_t\}$ and the last $k$ previous output set $\mathcal{O} = \{O_1,\ldots,O_t\}$ (Line~\ref{algline:for-context-builder}-\ref{algline:context-builder}).
Finally, we generate the rewritten question using the model $\mathcal{M}$ conditioned on the application specific prompt $P$, the context $C$ we constructed, and the current question $Q$ to obtain the rewritten question $\hat{Q} = \mathcal{M}(P,C,Q)$. 
As we will show, this general framework can recover many different approaches that are useful for a variety of tasks.

\subsection{Query Rewrite}\label{sec:query-rewrite}
In this section, we detail the query rewrite approach that we investigate in this work using Algorithm~\ref{alg:query-rewrite}.
This approach was designed with the question-answering task in mind
Notably, we of course utilize a custom application-specific prompt $P$ as input, that is different from the query fusion prompt used below.
This prompt consists of instructions to rewrite the input query and has application-specific aspects.
For the results reported later, we set $k=5$ and utilize both the set of previous input queries and set of previous responses.
As an aside, we denote this more general approach as ``query rewrite'', and refer to the other described next in Section~\ref{sec:query-fusion} as query fusion.

\begin{algorithm}[h!]
\caption{Parameterized Query Rewrite}
\label{alg:query-rewrite}
\begin{algorithmic}[1]
\State {\bf Input:}\\
\quad current query $Q$ \\
\quad application specific prompt $P$ \\
\quad chat history length $k$ \\
\quad previous input query set $\,\mathcal{I} = \{I_1, \ldots, I_t\}$\\
\quad previous response set $\mathcal{O} = \{O_1, \ldots, O_t\}$ \\
\quad base large language model $\mathcal{M}$

\vspace{1mm}
\State {\bf Output:} a rewritten query $\hat{Q}$
\vspace{2mm}

\State $C = \emptyset$ and $t = |\mathcal{I}|$ \label{algline:init}
\For{$i=\max(1, t-k)$ to $t$} \label{algline:for-context-builder}
\State $C \leftarrow C \cup \{I_{i}, O_{i}\}$ \label{algline:context-builder}
\EndFor

\State $\hat{Q} = \mathcal{M}(P, C, Q)$
\end{algorithmic}
\end{algorithm}

\subsection{Query Fusion}\label{sec:query-fusion}
Now, we describe the query fusion approach.
To obtain the query fusion approach from the parameterized approach in Algorithm~\ref{alg:query-rewrite},
we simply set $k=1$, then let $O_i=\emptyset$ for all $i=1,\ldots,|\mathcal{O}|$, and set $\mathcal{I} \leftarrow \mathcal{R}$ to be the set of previously rewritten queries $\mathcal{R} = \{R_1,\ldots,R_{t}\}$. 
An important property of the proposed query fusion is that even when $k=1$, this approach can succinctly capture an arbitrary number of previous questions, not only the last question or $k$.
This is a limitation of the approach described in Section~\ref{sec:query-rewrite} since it is highly dependent on selecting a good $k$, and this is a fundamentally challenging problem in itself, and depends on the user and application task.
An intuitive example is provided in Table~\ref{table:conv-and-rewritten-question-query-fusion}.
It is straightforward to see this property, for instance, our query fusion approach always leverages the last rewritten question as the $k=1$ previous question, and then fuses it with the current output, hence, suppose the last rewritten question is ``compare yearly revenue by country'', then the user asks ``show it as a line chart'', then the rewritten question is simply, ``compare yearly revenue by country as line chart''.
From Table~\ref{table:conv-and-rewritten-question-query-fusion}, it is straightforward to see that even after the user has asked 10 questions, the generated rewritten questions always contain much more information, and can be thought of as a compact summary of the previous conversation.

As an aside, this approach is likely to be useful for conversational text-to-image generation applications as well, since the typical user behavior is that a user requests to generate an image, such as ``generate an image of a cheetah'', then modifies the generated image such as ``add a tree on the right'', etc.
This pattern follows the same as the text-to-visualization application, among others that deal with modalities other than text.

The proposed parameterized approach provided in Algorithm~\ref{alg:query-rewrite} is able to recover both the query rewrite and query fusion approaches described above, as well as many others that are likely to be important for other applications.

\begin{table*}[h!]
    \centering
    \small
    \setlength{\tabcolsep}{10pt} %
    \renewcommand{\arraystretch}{1.2} %
    \caption{Summary of the datasets and their statistics}
    \vspace{-1mm}
    \begin{tabular}{>{\raggedright\arraybackslash}m{3.2cm} 
                    >{\centering\arraybackslash}m{1.55cm} 
                    >{\centering\arraybackslash}m{2.2cm} 
                    >{\centering\arraybackslash}m{1.2cm}
                    >{\centering\arraybackslash}m{1.2cm}
                    }
                    
                    \toprule
        & 
        \textcolor{black}{\textbf{\# Questions}} & 
        \textcolor{black}{\textbf{\# Questions with Chat History}} & 
        \textcolor{black}{
        \textbf{Chat Length}
        } &
        \textcolor{black}{\textbf{Question Types}} %
        \\
\midrule
\TTT
        \textbf{Text-based Q\&A} 
        & 179  & 136 & 2-5 & 3 \\
        \midrule
\TTT
        \textbf{Text-to-Vis (long conv.)}
        & 794  & 715 & 10 & 7 \\
        \textbf{Text-to-Vis (short conv.)}
        & 171  & 161 & 2 & 3\\

\bottomrule
    \end{tabular}
    \label{tab:data-stats}
\end{table*}

\definecolor{googleblue}{RGB}{66,133,244}
\definecolor{googlered}{RGB}{219,68,55}
\definecolor{googlegreen}{RGB}{15,157,88}
\definecolor{googlepurple}{RGB}{138,43,226}
\definecolor{lightred}{RGB}{255,198,196}
\definecolor{lightblue}{RGB}{197, 241, 255}
\definecolor{lightgreen}{RGB}{200, 247, 200}
\definecolor{lightpurple}{RGB}{230,230,250}
\definecolor{lightyellow}{RGB}{242, 232, 99}
\definecolor{lighterblue}{RGB}{197, 220, 255}
\definecolor{lighterred}{RGB}{253, 249, 205}
\definecolor{lightyellow}{RGB}{207, 161, 13}
\definecolor{darkbluenew}{rgb}{0,0.08,180}

\section{Methodology}
In this section, we provide details on the methodology including the different datasets investigated and how they were collected as well as the evaluation metrics used.
A summary of the dataset statistics and properties are provided in Table~\ref{tab:data-stats}.
Notably, we detail the total number of questions for each, the number of questions with chat history, and the average chat length.

\subsection{Text-based Q\&A Dataset}
In this dataset, we collected questions that users asked to an actual assistant developed to aid users in understanding how to use a specific product and its features, or answer questions about the datasets they had, which can be viewed as metadata questions that ask about their data.
The questions users asked that are related to a product 
are closely aligned with questions asked about documentation or documents related to it.
In this task, there were 22 annotators that were domain experts.
These experts were assigned a set of conversations and asked to provide the ground-truth rewritten query after each interaction.
The conversations varied greatly in the length, going from only 2 questions to conversations with 5 questions.
One example of a conversation might be "what is streaming segmentation", followed by "how does it differ from batch segmentation?".
Overall, the dataset consists of 179 questions in total, with 136 of them having chat history.

\subsection{Text-to-Visualization Datasets}
For the text-to-visualization task, we created two datasets for evaluation. 
These include a short conversation and a long conversation dataset.
We detail each of these datasets carefully below.

\subsubsection{Short Conversational Dataset}
In a completely different annotation task, we asked 12 domain experts to write an analytics question, then a likely follow-up question that may be incomplete, along with the rewritten ground-truth question that takes into account both. 
Furthermore, we asked the annotator to mark the rewritten question as N/A if a rewrite was not needed. 
Finally, we also asked the annotator to mark the intent of the rewritten query.
The annotators were told to focus on writing mostly questions that require rewriting, as those that do not are easy to obtain.
This provided us with a dataset of 171 short conversations, where 161 of them were detected as actually needing to be rewritten.
One example short conversation is:
\vspace{1mm}

\centerline{\textcolor{googlered}{\texttt{Compare orders by country}}} 

\vspace{1mm}
\noindent 
then the follow-up input question is:
\vspace{1mm}

\centerline{\textcolor{googleblue}{\texttt{Show top-5 countries as bar chart}}}

\vspace{1mm}
\noindent
finally, the ground-truth rewritten question is:
\vspace{1mm}

\centerline{\texttt{\textcolor{googlered}{Compare orders by} \textcolor{googleblue}{top-5} \textcolor{googlered}{countries} \textcolor{googleblue}{as bar}}}
\vspace{1mm}

\noindent
where the ground-truth rewritten question is color-coded by the different questions, and the text shown in blue in the final above rewritten question indicates the new information included.

\subsubsection{Long Conversational Dataset}
To collect this dataset, we asked 18 domain experts to complete as many conversations as they can with an AI assistant designed for analytics. We gave them background into the assistants capabilities. Furthermore, we asked them to write conversations that are typically at least 10 questions, and gave them a concrete example.
We guided them to write conversations that include a sequence of rewrites and mentioned that some conversations can have multiple topics indicating that for a conversation of length 10, queries 1-4 can be on related topic (requiring maintaining the context), while questions 5-10 can be a completely different topic, so question 5 in that case wouldn't require a rewrite.
We ask the experts to provide the ground-truth rewritten question for each step in the conversation, and provide the intent of the rewritten question.

This led us to collect a total of 794 questions, with 715 of them having chat history.
For one example conversation, see Table~\ref{table:conv-and-rewritten-question-query-fusion}.
In particular, we see a typical conversation that a user may ask. 
This conversation consists of 10 data analysis questions where the task is to generate a visualization.

\begin{table*}[h!]
    \centering
    \small
    \setlength{\tabcolsep}{12pt} %
    \renewcommand{\arraystretch}{1.2} %
    \caption{Results comparing rewriting and fusion approaches across two conversational tasks.
    }
    \vspace{-1mm}
    \begin{tabular}{>{\raggedright\arraybackslash}m{3.1cm} 
                    >{\raggedright\arraybackslash}m{2.5cm} 
                    >{\centering\arraybackslash}m{2.5cm} 
                    >{\centering\arraybackslash}m{2cm}}
                    \toprule
        \textcolor{black}{\textbf{Task}} & \textcolor{black}{\textbf{Approach}} & \textcolor{black}{\textbf{Cosine Similarity}} & \textcolor{black}{\textbf{BERT F1}} \\
\bottomrule
\rowcolor{lightblue}
\TTT        
        \textbf{Text-based Q\&A} & Query Fusion  & 0.826 & 0.751 \\
        \rowcolor{lightblue}
        & Query Rewrite  & \textbf{0.859} & \textbf{0.828} \\
        \hline
\rowcolor{lightgreen}
\TTT        
        \textbf{Text-to-Vis (long conv.)}
        & Query Fusion  & \textbf{0.820} & \textbf{0.773} \\
        \rowcolor{lightgreen}
        & Query Rewrite  & 0.760 & 0.734 \\

\noalign{\hrule height 0.7pt}
    \end{tabular}
    \label{tab:query_performance}
\end{table*}

\subsection{Evaluation Metrics}
For the experimental results in Section~\ref{sec:exp}, we use two evaluation metrics, namely, cosine similarity and BERT F1 score.
To compute the metrics, we use the same embedding model to ensure all results are completely consistent.
To derive cosine similarity, we obtain the embedding of the ground-truth rewritten question denoted as $\vy$ and then derive the embedding of the rewritten question from one of the approaches $\hat{\vy}$, then compute the cosine similairity between the actual $\vy$ and generated $\hat{\vy}$ as:
\begin{equation}
\mathbb{E}(\vy, \hat{\vy}) = \frac{\vy \cdot \hat{\vy}}{\|\vy\| \|\hat{\vy}\|}
\end{equation}
where $\mathbb{E}(\vy, \hat{\vy})=1$ if the two embeddings of the generated and actual rewritten text are identical and $\mathbb{E}(\vy, \hat{\vy})=0$ if the two embeddings are orthogonal implying no similarity.
We also compute the BERT F1 score for completeness.

\section{Experiments}\label{sec:exp}
In these experiments, we investigate the query rewrite and fusion approaches across two different conversational tasks including 
conversational text-based Q\&A (Section~\ref{sec:exp-conv-text-based-QA}) and 
conversational data analysis (Section~\ref{sec:exp-conv-data-analysis}).

\subsection{Conversational Text-based Q\&A Results} \label{sec:exp-conv-text-based-QA}
For this experiment, we study query rewrite and query fusion for conversational text-based Q\&A. 
Results are provided in \Cref{tab:query_performance}.
Notably, we see that query rewrite achieves better cosine similarity and BERT F1 scores compared to query fusion.
In particular, query rewrite achieves a mean relative gain of 3.9\% and 9.8\% over query fusion for cosine similarity and BERT F1 score, respectively (\Cref{tab:query_performance}).
This indicates the importance of considering both the previous questions along with the answers to those questions during the rewriting process.
In comparison, query fusion does not leverage the previous responses from the model.
Furthermore, query fusion also recursively fuses the past questions as long as it makes sense to do so, and thus, can adaptively consider both short and longer conversational contexts during the query rewriting process.
However, for the conversational text-based Q\&A task, there is often a need to include the answers (responses from the model) as well, as the current question the user asks can often be the result of a series of other questions and their answers, building up to the next question, and so on.
For instance, a user may ask how to use some feature of a product, then get a response, then the next question may use or require information from both the previous question, but more importantly the previous answer.
One other situation we noticed is that a user may ask a few questions, then ask another question which may not have been what they wanted or useful, and then afterwards ask another question that was closely related to the questions before the last.
In both situations, query fusion did not perform well since it doesn't consider previous answers as input, and does not handle gaps in the conversational context where a user may wish to return to a previous part of the conversation after asking a question that gave an unintended response.

\subsection{Conversational Data Analysis Results}\label{sec:exp-conv-data-analysis}
Now we investigate query rewrite (Sec.~\ref{sec:query-rewrite}) and query fusion (Sec.~\ref{sec:query-fusion}) for the conversational data analysis task.
In this task, the user often creates one or more visualizations to help them better understand and analyze their data.
One example conversation is shown in \Cref{table:conv-and-rewritten-question-query-fusion}.
For this task, we first investigate the long conversation text-to-vis dataset.
Results are shown in \Cref{tab:query_performance}.
In particular, the query fusion approach from our parameterized framework (Alg.~\ref{alg:query-rewrite}) achieves significantly better cosine similarity and BERT F1 score compared to query rewrite.
Query fusion achieves a mean relative gain of 7.6\% and 5.2\% over query rewrite for cosine similarity and BERT F1 score, respectively (\Cref{tab:query_performance}).
This indicates the importance of query fusion for the conversational data analysis task, as it can better summarize the conversations that often involve creating a visualization for some user-specific dataset, and then one or more queries that involve different changes to it, which may include adding, removing, or replacing a data attribute, time range, chart type, filter, among others.
For such data analysis conversations, these results indicate that the query rewrite approach that includes the previous $k=5$ questions may often not be very useful, as it may miss the important context in the conversation, introducing ambiguity, leading to poor performance.
In comparison, the query fusion uses $k=1$, but instead of the previous question, it leverages the previous rewritten question, which is essentially a summary of the relevant previous questions.

\begin{table}[t!]
    \centering
    \small
    \setlength{\tabcolsep}{12pt} %
    \renewcommand{\arraystretch}{1.2} %
    \caption{Results comparing rewriting and fusion approaches for short text-to-vis conversational task.}
    \vspace{-1mm}
    \begin{tabular}{%
    H
                    >{\raggedright\arraybackslash}m{2cm} 
                    >{\centering\arraybackslash}m{1.6cm} 
                    >{\centering\arraybackslash}m{1.6cm}
                    }
        \toprule
        \textcolor{black}{\textbf{Task}} & \textcolor{black}{\textbf{Approach}} & \textcolor{black}{\textbf{Cosine Sim.}} & \textcolor{black}{\textbf{BERT F1}} \\
        \midrule
        \textbf{Text-to-Vis} & Query Fusion & \textbf{0.925} & \textbf{0.856} \\
        & Query Rewrite & 0.857 & 0.837 \\
        \bottomrule
    \end{tabular}
    \label{tab:query_performance_short_conv}
\end{table}

As shown in \Cref{table:conv-and-rewritten-question-query-fusion}, the resulting rewritten question after every step represents a succinct and compact summary of the conversation up to that point in time.
For instance, the last input to the conversation in \Cref{table:conv-and-rewritten-question-query-fusion} is "\textrm{add revenue}", and the proposed query fusion approach from Sec.~\ref{sec:query-fusion} leverages "\textrm{add revenue}" along with the \emph{previous rewritten question} "\textrm{compare this month pageviews by top-5 marketing channels as bar}", and the model outputs 
"\textrm{compare this month pageviews and revenue by top-5 marketing channels as bar}", which represents the new compact summary of the conversation up to this point in time.
Hence, query fusion can naturally handle conversations of an arbitrary length, whereas in query rewrite the user must specify the length $k$ of the chat history to consider, which is fundamentally challenging and prone to errors.
Consider the previous example used above, the query rewrite approach would only be able to leverage the previous $k=5$ questions as input to generate a correct and useful rewritten question.
However, it is obvious that the rewritten question generated from the query rewrite approach that uses only the chat history from rows 5-9 in \Cref{table:conv-and-rewritten-question-query-fusion} would be incorrect, as it is missing much of the previous context from earlier in the chat, \eg, the last $k=5$ questions the user asked did not include anything about \texttt{marketing channel}, and thus is missing a fundamental piece of information needed to correctly answer the users question without completely losing context.
As an aside, the query rewrite approach may also struggle from conflicting user inputs, such as "\textrm{show top-3}" and "\textrm{what about top-5}" as the model may incorrectly rewrite based on the one in the more distant past rather than the more recent request, whereas the query fusion approach naturally handles such cases as it recursively generates a rewritten question at each step that compactly summarizes the entire conversation up to any point in time.

We also investigated shorter conversations for the data analysis task as shown in Table~\ref{tab:query_performance_short_conv}.
For this dataset, we observe similar results as before.
Notably, the query fusion approach achieves better performance across both metrics compared to the query rewrite, which is consistent with the previous findings from the long conversation dataset.

\subsection{Additional Results with Rewrite Classifier}

To analyze further the results, we investigated the impact of the query rewrite approach when using a rewrite classifier first, which inferred whether a query required rewriting or not.
Results are reported in Table~\ref{tab:query-rewrite-vs-fusion-with-rewrite-classifier}.
Notably, we see that performance improves slightly in Table~\ref{tab:query-rewrite-vs-fusion-with-rewrite-classifier} when compared to the previous results in Table~\ref{tab:query_performance}.

\begin{table}[h!]
    \centering
    \small
    \vspace{1mm}
    \setlength{\tabcolsep}{12pt} %
    \renewcommand{\arraystretch}{1.3} %
    \caption{Results comparing rewriting with an ambiguity detection classifier.
    }
    \vspace{-2mm}
    \begin{tabular}{>{\raggedright\arraybackslash}m{3.1cm} @{}
    H@{}
                    >{\centering\arraybackslash}m{1.6cm} 
                    >{\centering\arraybackslash}m{1.3cm}}
                    \toprule
        \textcolor{black}{\textbf{Task}} & \textcolor{black}{\textbf{Approach}} & \textcolor{black}{\textbf{Cosine Sim.}} & \textcolor{black}{\textbf{BERT F1}} \\
        \bottomrule
        \textbf{Text-based Q\&A} 
        & Query Rewrite  & \text{0.871} & \text{0.859} \\
        \textbf{Text-to-Vis (long conv.)}
        & Query Rewrite  & 0.769 & 0.740 \\
        \hline
    \end{tabular}
    \label{tab:query-rewrite-vs-fusion-with-rewrite-classifier}
\end{table}

\section{Conclusion} \label{sec:conc}
In this work, we proposed a parameterized query rewrite approach that gives rise to a family of algorithms that can be used for a variety of different applications.
We investigated two such algorithms from the framework, and showed how this parameterized algorithm is able to recover both at either extreme.
We systematically compared these two different approaches, which we denoted as rewriting and fusion, on two fundamentally different generation tasks, including a text-to-text generation task and a multimodal generative task that takes as input text and generates a visualization or data table that answers the user's question.
Notably, we make several important findings.
Our results showed that the specific rewriting or fusion approach highly depends on the underlying use case and generative task.
In particular, we find that for a conversational question-answering assistant, the query rewriting approach performs best, whereas for a data analysis assistant that generates visualizations and data tables based on the user's conversation with the assistant, the fusion approach works best.

\bibliography{main}
\bibliographystyle{acl_natbib}

\end{document}